\documentclass[10pt,letterpaper]{uconn-minds}

\usepackage{amsmath,amssymb}
\usepackage{array}
\usepackage{tikz}
\usetikzlibrary{positioning,fit}
\usepackage{algorithm}
\usepackage{algpseudocode}
\usepackage{xspace}
\definecolor{promptblue}{HTML}{2F5D8C}
\definecolor{promptbg}{HTML}{F5F8FC}
\newsavebox{\promptcontent}
\newlength{\promptinner}
\newenvironment{promptbox}[1]
  {\def\prompttitle{#1}%
   \setlength{\promptinner}{\dimexpr\linewidth-18pt\relax}%
   \begin{lrbox}{\promptcontent}\begin{minipage}{\promptinner}\footnotesize
   \setlength{\parskip}{2pt}\setlength{\parindent}{0pt}}
  {\end{minipage}\end{lrbox}%
   \par\vspace{4pt}\noindent\begin{tikzpicture}
     \node[fill=promptbg, draw=promptblue, line width=0.8pt, rounded corners=2pt, inner xsep=8.6pt,
           inner ysep=5pt, anchor=north west] (body) at (0,0) {\usebox{\promptcontent}};
     \node[fill=promptblue, draw=promptblue, line width=0.8pt, rounded corners=2pt, anchor=south west,
           inner xsep=8.6pt, inner ysep=4pt, text width=\promptinner, text=white,
           font=\sffamily\bfseries\small] at (body.north west) {\prompttitle};
   \end{tikzpicture}\par\vspace{2pt}}
\newcommand{\ph}[1]{{\color{promptblue}\texttt{\{#1\}}}}
\newenvironment{promptlist}
  {\begin{list}{\textbullet}{\setlength{\leftmargin}{12pt}\setlength{\labelwidth}{8pt}\setlength{\labelsep}{4pt}%
   \setlength{\itemsep}{1pt}\setlength{\parsep}{0pt}\setlength{\topsep}{2pt}}}
  {\end{list}}

\newcommand{\methodname}{BLADE\xspace}
\newcommand{\methodfull}{\underline{B}lame \underline{L}ocalization via \underline{A}udited \underline{D}ependency \underline{E}vidence}
\title{\methodname: Auditing Information Use for Failure Attribution\\in LLM Multi-Agent Systems}
\author{Xiaoguang Guo, Yaohao Long, Chuxu Zhang$^{\dag}$\\
University of Connecticut \\
\texttt{\{xiaoguang.guo, chuxu.zhang\}@uconn.edu}, \texttt{robinlong5678@gmail.com}\\
{\textsuperscript{\dag}Corresponding author}}
\runningtitle{BLADE: Failure Attribution in LLM Multi-Agent Systems}
\sourcecode{https://github.com/GXG-CS/BLADE}
\hypersetup{
  pdftitle={BLADE: Auditing Information Use for Failure Attribution in LLM Multi-Agent Systems},
  pdfauthor={Xiaoguang Guo, Yaohao Long, and Chuxu Zhang},
  pdfsubject={Research preprint},
  pdfkeywords={failure attribution, LLM multi-agent systems, agent debugging}
}
\paperabstract{%
Large language model multi-agent systems extend individual agents'
capabilities through collaboration across specialized roles, tools, and
workflows. Yet collaboration makes failures harder to diagnose: one agent may
misinterpret a correct result while others carry the mistake forward. Failure
attribution therefore seeks to identify the responsible agent and decisive error step.
Recent methods trace dependencies, but knowing where information came from is
not the same as knowing where it went wrong. We introduce \methodname
(\methodfull), a training-free framework that moves beyond tracing information
to auditing its use. Using Jev for structured
probabilistic judgments, \methodname uses dependency probabilities to select
relevant \mbox{source--decision} handoffs for auditing. At these handoffs, it
examines whether receiving decisions use their sources faithfully and within
task requirements, distinguishing error introduction, propagation, and repair.
These findings inform a global diagnosis that considers all execution steps.
Without reference answers, \methodname achieves the best or tied-best
step accuracy among eight strategies under a shared Jev decision interface on
1,228 failed runs across five Who\&When and Who\&When Pro subsets,
outperforming All-at-once by 2.2--11.1 percentage points. Making responsibility
evidence explicit offers a path toward more interpretable and reliable
agentic systems.}

\begin{document}
\maketitle
\thispagestyle{firstpage}

\section{Introduction}
\label{sec:introduction}
Large language model (LLM)-based agents are increasingly being developed to
tackle complex tasks through sustained reasoning, tool use, and multi-agent
collaboration \citep{fourney2024magentic,luo2026ultrahorizon}. These advances
are enabling increasingly sophisticated workflows in software engineering,
web-based tasks, and scientific discovery
\citep{hong2024metagpt,ghareeb2026scientific}. However, as these systems
undertake more complex and interdependent tasks, diagnosing their failures
becomes increasingly challenging. Errors arising from incorrect reasoning,
misinterpreted observations, or faulty information exchange may propagate
across multiple agents before surfacing in the final outcome
\citep{cemri2025masfail,lin2026agentask}. Consequently, observing a failed
outcome, or identifying erroneous intermediate decisions, does not
necessarily reveal which mistakes were responsible for the failure
\citep{zhang2025whoandwhen,qi2026trajdebug}. This challenge motivates the study
of failure attribution, which seeks to locate decisive errors and assign
responsibility within execution trajectories, providing a foundation for
debugging, corrective feedback, and improving the reliability of
agentic systems \citep{zhang2025whoandwhen,chen2026traceelephant}.

\begin{figure}[t]
  \centering
  \includegraphics[width=0.5\linewidth]{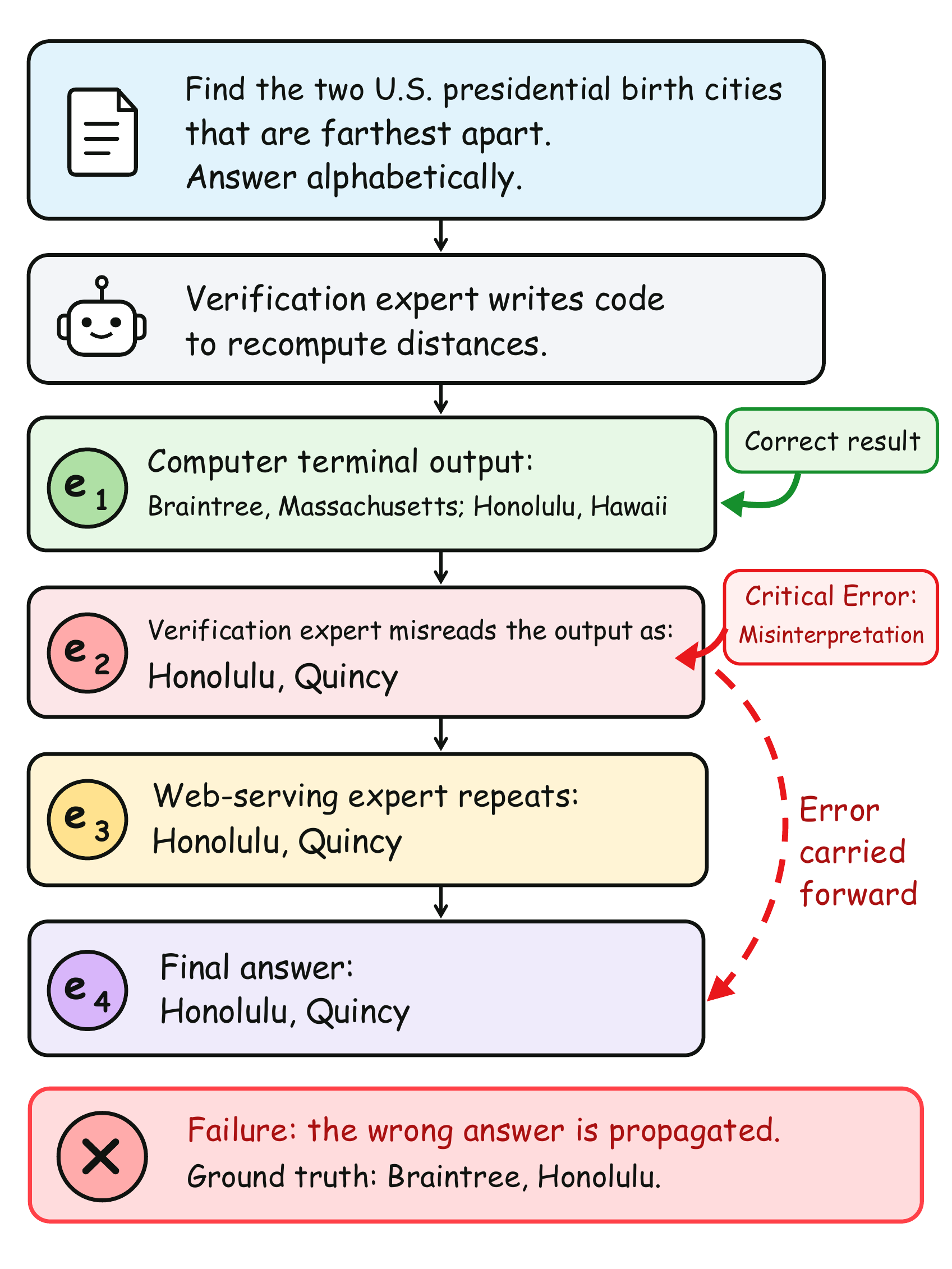}
  \caption{Failure attribution requires distinguishing where an error is
    first introduced from where it is merely propagated downstream.}
  \label{fig:handoff-case}
\end{figure}

Despite its importance, failure attribution in multi-agent systems presents
three closely related challenges. First, identifying relevant interactions
is difficult, as critical decisions may depend on information from distant
steps and multiple agents, making it difficult to determine which
dependencies deserve closer attention \citep{rafi2026falat,deng2026defa}.
Second, tracing dependencies alone is not enough to establish responsibility.
An agent may faithfully pass along incorrect information or introduce a new
error by misinterpreting a correct source
\citep{cemri2025masfail,zhang2025whoandwhen}. For example, as shown in
Figure~\ref{fig:handoff-case}, a code execution correctly identifies Braintree and
Honolulu as the farthest-apart birthplaces of U.S.\ presidents, but a
subsequent agent misreads the result as Honolulu and Quincy, which is then
carried into the final answer. The error therefore arises not from the
original source, but from how its information is interpreted. Third,
identifying local errors does not necessarily lead to a correct overall
attribution, since some mistakes may be corrected later, while others may
have little impact on the final outcome \citep{qi2026trajdebug,rafi2026falat}.
Together, these challenges call for an approach that goes beyond tracing
dependencies to examine how agents use information and how local evidence
contributes to the overall diagnosis.

Recent studies have explored several approaches to failure attribution.
Counterfactual replay has been used to construct training supervision for
attribution models \citep{zhang2026agentracer}, while repeated-run analyses
assess the contributions of individual agent actions \citep{ge2026famas}.
Dependency-guided methods trace information flow and error propagation across
execution steps \citep{rafi2026falat,deng2026defa}. Other approaches examine
how errors evolve over time, verify failure hypotheses, or check for
violations of task and tool requirements
\citep{qi2026trajdebug,li2026errorprobe,barke2026agentrx}. These advances have
improved our understanding of how failures develop. Building on this progress,
we ask how information use at source--decision handoffs can provide explicit
evidence of responsibility. Beyond identifying which steps are connected,
we argue that attribution should examine what an agent receives, how it uses
that information, and whether its decisions satisfy the task's requirements.
This motivates treating each source--decision handoff as a unit for collecting
evidence of error introduction, propagation, and repair. Since these
interactions can be numerous and their assessment uncertain, such analysis
calls for selective, structured judgments over a shared execution context,
with their probabilities informing the final diagnosis.

To realize this perspective, we introduce \methodname (\methodfull), a
training-free framework that moves beyond tracing information flow to
examining responsibility at agent handoffs. At its core, \methodname extends
a dependency graph into a blame graph: rather than merely recording which
decisions rely on earlier information, it also captures how that information
is used and where errors may be introduced, propagated, or repaired.
\methodname first reconstructs relationships among execution steps and uses
their probabilities to select relevant source--decision handoffs for closer
examination. Inspired by contract-based blame assignment
\citep{findler2002contracts,wadler2009welltyped}, it then audits how each
receiving decision interprets and applies its source under the task's
requirements, distinguishing newly introduced violations from inherited
errors and repairs. The resulting findings are associated with the audited
handoffs as structured evidence rather than fixed blame assignments.
\methodname implements these judgments with Jev, whose decision interface
evaluates independent structured questions in parallel over a shared
execution context and returns probabilities over predefined alternatives.
Finally, \methodname combines the blame graph with the complete trajectory
to identify the decisive error and assign responsibility, keeping every
execution step as a candidate and allowing the global diagnosis to override
uncertain local findings.

We evaluate \methodname on 1,228 failed executions across five subsets of
Who\&When and Who\&When Pro, covering diverse multi-agent architectures and
both natural and injected failures. Under a shared Jev decision interface,
\methodname achieves the best or tied-best step attribution accuracy among
eight compared strategies on every subset. Our ablations support the central
distinction between tracing information and auditing its use: dependency
information alone yields mixed improvements, whereas adding handoff-level
audit findings improves step accuracy by 2.4--3.4 percentage points across
all four subsets examined. Shuffling dependency edges and audit findings
across event pairs reduces accuracy, highlighting the importance of
grounding evidence in the interactions it describes. Dependency-guided
auditing matches or outperforms exhaustive auditing on three subsets
while examining only 13--23\% as many handoffs. These results
demonstrate the value of using dependencies to guide targeted audits and
the evidence to support global attribution.

We make the following contributions:
\begin{itemize}
  \item \textbf{Beyond Dependency Tracing.} We identify source--decision
    handoffs as units for collecting responsibility evidence, distinguishing
    information provenance from how that information is used. Dependency
    tracing reveals how information flows between agents, but not whether
    an error originates in a receiving decision or is inherited from an
    upstream source. We argue that source--decision handoffs are therefore
    critical to understanding how failures arise and propagate.
  \item \textbf{\methodname Framework.} We introduce \methodname, a
    training-free framework that uses probabilistic dependencies to guide
    where to inspect rather than directly assigning blame. By auditing how
    agents interpret and use information, \methodname constructs a blame
    graph capturing error introduction, propagation, and repair. This
    evidence informs a global diagnosis in which all execution steps remain
    candidates.
  \item \textbf{Empirical Insights.} Experiments on 1,228 failed trajectories
    across five benchmark subsets reveal that dependency information alone
    offers inconsistent benefits, whereas handoff auditing consistently
    improves attribution across the four subsets examined in our ablations.
    Selective auditing also matches or outperforms exhaustive auditing on
    three subsets, showing that inspecting more interactions does not
    necessarily improve attribution.
\end{itemize}

\section{Related Work}
\label{sec:related-work}

\subsection{Failure Attribution in Agent Trajectories}
Who\&When \citep{zhang2025whoandwhen} established agent- and step-level
failure attribution, with subsequent benchmarks covering injected failures
\citep{liu2026pro} and richer execution traces
\citep{chen2026traceelephant}. Beyond direct trajectory judgments, AgenTracer
\citep{zhang2026agentracer} trains an attribution model on supervision derived
from counterfactual replay, while FAMAS \citep{ge2026famas} estimates
responsibility across repeated executions. These approaches obtain
additional attribution signals through training or repeated runs, motivating
complementary forms of diagnosis.

Other methods focus on collecting evidence of failure. AgentRx
\citep{barke2026agentrx} checks task and tool constraints, ErrorProbe
\citep{li2026errorprobe} verifies failure hypotheses through tool-grounded
execution, and TrajDebug \citep{qi2026trajdebug} tracks errors through their
correction and downstream impact. Together, they highlight an important
distinction: detecting an error is not equivalent to establishing
responsibility. In collaborative workflows, this often depends on how agents
interpret and use information produced by others.

\subsection{From Dependency Tracing to Responsibility Evidence}
Dependency-aware methods examine these cross-step interactions. CHIEF
\citep{wang2026chief} and DCFA \citep{wang2026dcfa} organize failures through
causal-inspired structures, while TRIAGE \citep{wang2026triage} ranks steps
by suspicion and structural impact. More closely related, FALAT
\citep{rafi2026falat} uses dependencies to distinguish error origins,
propagation, and correction, whereas DeFA \citep{deng2026defa} combines event
dependencies with failure propagation analysis. These methods demonstrate
the value of structural evidence, yet a dependency does not itself establish
whether the receiving decision used its source correctly. Moreover,
uncertain local evidence raises the question of which interactions warrant
closer examination and how strongly they should influence attribution.

This distinction echoes data provenance and program slicing, which trace
information dependencies \citep{buneman2001why,weiser1981slicing}, and
contract-based blame assignment, which examines responsibility at
interaction boundaries \citep{findler2002contracts}. \methodname builds on
these ideas by using dependency probabilities to select source--decision
handoffs for targeted auditing. Its typed findings capture how information
is used, distinguishing error introduction from propagation and repair.
Unlike candidate-focused adjudication in DeFA, these findings do not restrict
the final search space: every step remains eligible for global attribution.
Dependencies thus guide where to inspect, while the resulting evidence
informs where responsibility lies.

\section{\methodname}
\label{sec:method}
\begin{figure}[t]
  \centering
  \includegraphics[width=\textwidth]{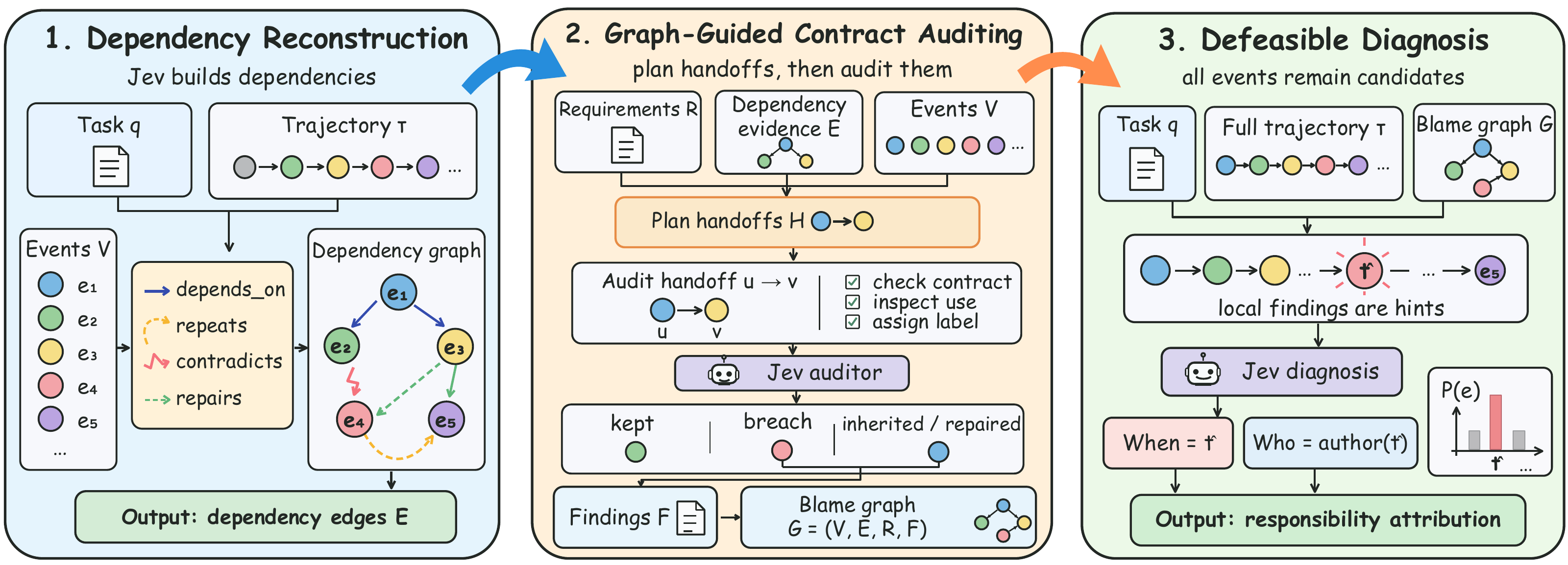}
  \caption{Overview of \methodname. (1) Jev reconstructs dependencies between
    events. (2) The dependencies select the handoffs to audit, and Jev checks how
    each agent decision uses its source against the task's requirement spans
    $R$. (3) The diagnosis reads the full trajectory with the blame
    graph $G=(V,E,R,F)$; every event remains a candidate.}
  \label{fig:overview}
\end{figure}
\methodname separates the construction of diagnostic evidence from the
final assignment of responsibility. Rather than using dependencies to
directly determine blame, it uses them to guide targeted inspection of
source--decision handoffs. These audits provide structured evidence
about how information is used, while the final diagnosis considers
the complete trajectory without restricting the candidate events
(Figure~\ref{fig:overview}, Algorithm~\ref{alg:blade}).
We implement these judgments with Jev, a decision model that returns
probabilities over predefined alternatives rather than free-form text.

\subsection{Problem Definition}
\label{sec:problem}
Following \citet{zhang2025whoandwhen}, consider a multi-agent system
attempting a task $q$. A failed execution produces a trajectory
$\tau=(e_1,\dots,e_T)$, where each event $e_t$ records a message,
action, or observation. Let $V=\{e_1,\dots,e_T\}$ denote the event
set, and let $a(e_t)$ denote the recorded actor identifier of event $e_t$,
which may be undefined ($\bot$).

Failure attribution aims to identify the responsible agent $a^*$
and decisive error step $t^*$. Following Who\&When, an error is
decisive if correcting it would turn the failed execution into a
successful one, with preceding steps unchanged and subsequent
execution adjusted accordingly. If multiple decisive errors exist,
the target is the earliest.

Given only $q$ and $\tau$, \methodname predicts
$\hat t\in\{1,\dots,T\}$ and assigns the recorded actor of the
selected event as responsible, $\hat a=a(e_{\hat t})$.
Neither the reference answer nor the ground-truth attribution is
available during inference.

\begin{algorithm}[t]
  \caption{\methodname: Dependency-guided responsibility attribution.}
  \label{alg:blade}
  \small
  \begin{algorithmic}[1]
    \Require Task $q$, trajectory $\tau=(e_1,\dots,e_T)$,
      threshold $\theta$, maximum sources per event $K$
    \Statex \textit{(1) Dependency reconstruction}
    \For{$t=2,\dots,T$, $r\in\mathcal{R}_{\mathrm{link}}$}
      \State $p^{(r)}_{ut}\gets P_{\mathrm{Jev}}(e_u\mid s_\tau,x_{t,r})$ for $u<t$
    \EndFor
    \State $E\gets\{(e_u,e_t,r,p^{(r)}_{ut}): p^{(r)}_{ut}\ge\theta\}$
    \Statex \textit{(2) Graph-guided contract auditing}
    \State $\mathcal{H}\gets\bigcup_t\mathcal{H}_t$ \Comment{Eq.~\ref{eq:audit-plan}}
    \State $R\gets$ requirement spans of $q$
    \State $\kappa(e_t)\gets$ decision/evidence classification via Jev
    \Statex \hspace{\algorithmicindent} for all $e_t\in V$
    \State $\ell_{ut},\pi_{ut}\gets$ Jev's semantic audit of $e_u\to e_t$
    \Statex \hspace{\algorithmicindent} for $(e_u,e_t)\in\mathcal{H}$, given $R$
    \State $F\gets$ non-kept findings whose receiving
    \Statex \hspace{\algorithmicindent} events are agent decisions \Comment{Eq.~\ref{eq:findings}}
    \State $G\gets(V,E,R,F)$
    \Statex \textit{(3) Defeasible diagnosis}
    \State $\hat t\gets\arg\max_t P_{\mathrm{Jev}}(e_t\mid q,\tau,G)$ \Comment{all events}
    \State \Return $\hat t,\;a(e_{\hat t})$
  \end{algorithmic}
\end{algorithm}

\subsection{Dependency Reconstruction}
\label{sec:dependency}
The first challenge in constructing responsibility evidence is
identifying which interactions warrant closer examination.
In multi-agent trajectories, a decision may rely on a distant
instruction, tool result, or intermediate claim rather than
the immediately preceding event. Program slicing and data
provenance have long studied how outputs depend on earlier
computations and information
\citep{weiser1981slicing,buneman2001why}.
In agent trajectories, however, many relevant dependencies are
implicit in natural-language interactions and must be inferred
from trajectory context.
\methodname reconstructs these relations as probabilistic evidence,
using them to guide targeted auditing rather than directly
determining responsibility.

Given a shared state $s_\tau$ containing the task and trajectory,
\methodname uses Jev (TypeSafe, \texttt{jev-1.13.0})%
\footnote{\url{https://docs.typesafe.ai}}
to assess four relations for each event $e_t$ with $t>1$:
\emph{depends on}, \emph{repeats}, \emph{contradicts}, and
\emph{repairs}. Each relation is evaluated through a separate
multiple-choice question asking which preceding event in
$\{e_1,\dots,e_{t-1}\}$ best matches the relation, with
\textsc{none} as an additional option. Jev returns probabilities
over these alternatives, with questions sharing the same context
evaluated independently and batched into requests
(Appendix~\ref{sec:app-questions}).

Let $\mathcal{R}_{\mathrm{link}}$ denote the four relation types and
$x_{t,r}$ the question for relation $r$ at event $e_t$.
The probability assigned to an earlier event $e_u$ is
\begin{equation}
  p_{ut}^{(r)}=P_{\mathrm{Jev}}(e_u\mid s_\tau,x_{t,r}).
\end{equation}
We retain relations with probability at least $\theta=0.3$,
forming the weighted edge set
\begin{equation}
  \begin{aligned}
    E=\bigl\{&(e_u,e_t,r,p_{ut}^{(r)}): u<t,\\
             &r\in\mathcal{R}_{\mathrm{link}},\ p_{ut}^{(r)}\ge\theta\bigr\}.
  \end{aligned}
\end{equation}

These edges represent plausible relations rather than verified
causal links. The relation types serve complementary roles:
the full \emph{depends-on} distributions guide the selection
of source--decision handoffs for auditing, while all four
relations provide structural context for global diagnosis.
Dependency reconstruction thus identifies where information
may have flowed; determining how it was used requires
examining the handoffs themselves.

\subsection{Graph-Guided Contract Auditing}
\label{sec:auditing}
An erroneous decision need not be the source of its own error.
The same incorrect claim may reflect a misinterpretation of a
correct source or faithful propagation of an upstream mistake.
Distinguishing these cases requires examining how the receiving
decision interprets and uses its source. \methodname therefore treats the
source--decision handoff as the unit of diagnostic evidence.
Drawing on contract-based blame assignment
\citep{findler2002contracts}, it examines whether receiving
decisions use their sources faithfully and in accordance
with the task's requirements.

\paragraph{Audit planning.}
Not all event pairs represent meaningful information
transfers. Exhaustively auditing them requires a quadratic
number of judgments and may introduce findings from
weakly related interactions. \methodname instead uses
the \emph{depends-on} distributions to prioritize plausible
handoffs. For each event $e_t$, let
$\mathcal{U}_t=\{u<t:p_{ut}^{(\mathrm{dep})}\ge\theta\}$, where
$\mathrm{dep}$ denotes the \emph{depends-on} relation, and let
$\mathcal{S}_t\subseteq\mathcal{U}_t$ be the indices of the at most $K=2$
sources with the highest $p_{ut}^{(\mathrm{dep})}$. The selected handoffs are
\begin{equation}
  \mathcal{H}_t=\bigl\{(e_u,e_t): u\in\mathcal{S}_t\bigr\},
  \label{eq:audit-plan}
\end{equation}
and $\mathcal{H}=\bigcup_{t=2}^{T}\mathcal{H}_t$.
Audit planning reuses the dependency probabilities and
requires no additional model calls.
Crucially, selecting what to inspect does not restrict
which events may ultimately be held responsible.

\paragraph{Semantic contract auditing.}
Each handoff $(e_u,e_t)$ carries an implicit semantic
contract: the receiving decision should preserve its
source's meaning, avoid treating uncertain information
as established fact, and respect the task's requirements.
Unlike formal contract verification, these obligations
are assessed through structured semantic judgments.
\methodname extracts requirement spans $R$ from the task $q$
(Appendix~\ref{sec:app-spans}) and includes
them with the complete trajectory in the audit state
$s_{\tau,R}$.

Jev classifies each event as an agent decision or evidence,
denoted by $\kappa(e_t)$, and evaluates how each selected
receiving event $e_t$ uses its source $e_u$. For question
$x_{ut}^{\mathrm{use}}$, Jev assigns probabilities over nine
predefined labels $\mathcal{L}$:
\begin{equation}
  p_{ut}(\ell)=P_{\mathrm{Jev}}\big(\ell\mid s_{\tau,R},x^{\mathrm{use}}_{ut}\big),\qquad \ell\in\mathcal{L}.
\end{equation}
Let $\ell_{ut}$ be the most probable label and
$\pi_{ut}$ its assigned probability.

\paragraph{Responsibility findings.}
The labels distinguish potential error introduction,
inheritance or repair of existing errors, and cases yielding
no retained finding. We organize them into three groups,
$\mathcal{L}=\mathcal{L}_{\mathrm{kept}}
\cup\mathcal{L}_{\mathrm{breach}}
\cup\mathcal{L}_{\mathrm{inh}}$:
\begin{itemize}
  \item \emph{Kept}: faithful use, no actual use of the source,
    or insufficient evidence to assess the handoff.
  \item \emph{Breach}: potential error introduction through
    unverified adoption, inapplicable information,
    misinterpretation, or a correction that introduces
    a new error.
  \item \emph{Inherited or repaired}: propagation of
    an existing error, or correction without introducing
    a new one.
\end{itemize}

For handoffs into agent decisions, \methodname retains
non-kept judgments as structured findings. Let
$\mathcal{H}^{\mathrm{dec}}=
\{(e_u,e_t)\in\mathcal{H}:
\kappa(e_t)=\mathrm{decision}\}$.
The findings are
\begin{equation}
  \begin{aligned}
    F=\bigl\{&(e_u,e_t,\ell_{ut},\pi_{ut}) :\\
    &(e_u,e_t)\in\mathcal{H}^{\mathrm{dec}},\ \ell_{ut}\notin\mathcal{L}_{\mathrm{kept}}\bigr\}.
  \end{aligned}
  \label{eq:findings}
\end{equation}

\paragraph{Blame graph.}
The resulting \emph{blame graph} $G=(V,E,R,F)$ connects
where information came from with how it was used,
combining dependency relations and handoff findings
as evidence for global attribution. Here, $E$ captures
plausible information relationships, $F$ records
information-use findings at selected handoffs, and
$R$ provides the task requirements against which
these interactions are assessed.

\subsection{Defeasible Diagnosis}
\label{sec:diagnosis}
Handoff-level evidence reveals how errors are introduced,
propagated, or repaired, but does not necessarily identify
the decisive error. A local breach may be corrected later,
have no effect on the outcome, or coexist with other errors.
Conversely, a decisive error may lack an audit finding.
Confidence in an information-use label does
not imply that the corresponding event is responsible for
the failure. Although each audit considers the complete
trajectory, it answers a handoff-specific question.
Global diagnosis must additionally determine which error
was decisive for the failed execution.

This distinction draws on model-based diagnosis, which
seeks explanations for observed system failures
\citep{reiter1987theory}, and truth maintenance, which
allows conclusions to be revised as their supporting
evidence changes \citep{doyle1979truth}.
Following this perspective, \methodname treats the
blame graph as revisable diagnostic evidence rather than
a fixed assignment of responsibility.

Given the task $q$, complete trajectory $\tau$, and
blame graph $G$, Jev answers the same decisive-step
question as the All-at-once baseline, augmented with
structured responsibility evidence:
\begin{equation}
  \begin{aligned}
    \hat t&=\arg\max_{t\in\{1,\dots,T\}}
      P_{\mathrm{Jev}}\bigl(e_t\mid q,\tau,G\bigr),\\
    \hat a&=a(e_{\hat t}).
  \end{aligned}
\end{equation}
Every event remains a candidate, including those without
audit findings, allowing the global judgment to reconsider
uncertain local evidence and distinguish error origins from
downstream symptoms. By separating evidence collection
from attribution, \methodname uses dependencies to guide
inspection, handoff audits to characterize information use,
and global diagnosis to identify the decisive error.

\section{Experiments}
\label{sec:experiments}

\begin{table}[t]
  \centering
  \small
  \setlength{\tabcolsep}{3.5pt}
  \begin{tabular}{@{}lcccccccccccc@{}}
    \toprule
    & \multicolumn{4}{c}{Who\&When} & \multicolumn{6}{c}{Who\&When Pro} & \multicolumn{2}{c}{Macro Avg.} \\
    \cmidrule(lr){2-5}\cmidrule(lr){6-11}\cmidrule(l){12-13}
    & \multicolumn{2}{c}{HC} & \multicolumn{2}{c}{AG}
    & \multicolumn{2}{c}{Magentic-One} & \multicolumn{2}{c}{DyLAN} & \multicolumn{2}{c}{MetaGPT} & \multicolumn{2}{c}{} \\
    \cmidrule(lr){2-3}\cmidrule(lr){4-5}\cmidrule(lr){6-7}\cmidrule(lr){8-9}\cmidrule(lr){10-11}
    Method & Agent & Step & Agent & Step & Agent & Step & Agent & Step & Agent & Step & Agent & Step \\
    \midrule
    All-at-once \citep{zhang2025whoandwhen} & 50.0 & 24.1 & \underline{64.3} & \underline{50.0} & 80.4 & 78.7 & \textbf{59.9} & \underline{52.6} & 46.9 & 54.3 & 60.3 & \underline{52.0} \\
    Step-by-step \citep{zhang2025whoandwhen} & 55.2 & \underline{27.6} & 57.9 & 39.7 & 80.5 & 76.0 & 50.7 & 46.1 & 53.1 & 53.1 & 59.5 & 48.5 \\
    Binary search \citep{zhang2025whoandwhen} & 56.9 & \textbf{31.0} & 59.5 & 45.2 & 74.6 & 68.7 & 58.6 & \textbf{56.6} & 37.0 & 37.0 & 57.3 & 47.7 \\
    \midrule
    TRIAGE \citep{wang2026triage} & 56.9 & 22.4 & 60.3 & 44.4 & 83.5 & 79.9 & 52.6 & 48.0 & 46.9 & 46.9 & 60.0 & 48.3 \\
    FALAT \citep{rafi2026falat} & 48.3 & 20.7 & 56.3 & 37.3 & 76.1 & 69.3 & 40.1 & 34.2 & 37.0 & 37.0 & 51.6 & 39.7 \\
    \midrule
    ErrorProbe \citep{li2026errorprobe} & \textbf{63.8} & 20.7 & 59.5 & 39.7 & 77.8 & 74.7 & 52.0 & 42.1 & 29.6 & 29.6 & 56.5 & 41.4 \\
    TrajDebug \citep{qi2026trajdebug} & \underline{58.6} & 22.4 & 62.7 & 47.6 & \underline{83.7} & \underline{80.0} & 52.0 & 42.8 & \underline{63.0} & \underline{63.0} & \underline{64.0} & 51.2 \\
    \midrule
    \methodname (ours) & \underline{58.6} & \textbf{31.0} & \textbf{66.7} & \textbf{54.0} & \textbf{84.0} & \textbf{80.9} & \underline{59.2} & \textbf{56.6} & \textbf{65.4} & \textbf{65.4} & \textbf{66.8} & \textbf{57.6} \\
    \bottomrule
  \end{tabular}
  \caption{Agent and step accuracy (\%) without the reference answer. All
    methods are instantiated under the same Jev decision interface
    (Appendix~\ref{sec:app-baselines}). HC: Hand-Crafted; AG: Algorithm-Generated.
    Macro Avg. equally weights the five subsets. Bold marks the best and
    underline the second best.}
  \label{tab:main}
\end{table}

\subsection{Experimental Setup}
\label{sec:setup}
\paragraph{Benchmarks.}
We evaluate \methodname on 1,228 failed execution trajectories from two
established benchmarks: \textbf{Who\&When} \citep{zhang2025whoandwhen},
featuring naturally occurring failures, and \textbf{Who\&When Pro}
\citep{liu2026pro}, featuring injected failures.
Who\&When includes the Hand-Crafted (HC) subset based on Magentic-One
\citep{fourney2024magentic} and the Algorithm-Generated (AG) subset built
using CaptainAgent \citep{song2024captainagent}.
Our evaluation on Who\&When Pro covers three distinct multi-agent
architectures: Magentic-One, the debate-based DyLAN \citep{liu2024dylan},
and the pipeline-based MetaGPT \citep{hong2024metagpt}.
Together, these five subsets cover diverse collaboration structures and
both natural and injected failure settings for evaluating attribution.

\paragraph{Baselines.}
We compare \methodname against seven representative failure attribution
strategies from three categories.
\textbf{Direct attribution} includes All-at-once, Step-by-step, and Binary
search \citep{zhang2025whoandwhen}, which identify decisive errors through
different trajectory inspection strategies.
\textbf{Structure-guided localization} includes TRIAGE \citep{wang2026triage}
and FALAT \citep{rafi2026falat}, which use structural and dependency
information for failure localization.
\textbf{Error-aware diagnosis} includes ErrorProbe \citep{li2026errorprobe}
and TrajDebug \citep{qi2026trajdebug}, which examine failure evidence and
error evolution.
For consistent evaluation, all strategies are instantiated under a shared
Jev decision interface using identical tasks and recorded trajectories.
Implementation details for each strategy are provided in Appendix~\ref{sec:app-baselines}.

\paragraph{Metrics and protocol.}
We report \textbf{agent accuracy} and \textbf{step accuracy}, measuring
agreement with the annotated responsible agent and decisive error step,
respectively.
For DyLAN, whose annotations specify an agent and a round, a step prediction
is correct if it identifies a turn of the annotated agent in that round.
All methods receive the same task and trajectory without access to reference
answers or ground-truth attribution labels.
We remove benchmark-construction metadata that reveals injected error
locations and apply a consistent context-length policy across methods
(Appendix~\ref{sec:app-inputs}).
Invalid or missing predictions are counted as incorrect.

\subsection{Attribution Performance}
\label{sec:main-results}
\paragraph{Overall attribution accuracy.}
Table~\ref{tab:main} reports the performance of \methodname and seven baselines
across five subsets. With each subset weighted equally, \methodname achieves
the highest macro-average accuracy for both step attribution (57.6\%) and
agent attribution (66.8\%). These exceed the strongest baseline averages by
5.6 percentage points for step attribution (All-at-once, 52.0\%) and
2.8 points for agent attribution (TrajDebug, 64.0\%).
\methodname also achieves the best or tied-best step accuracy on every
subset, with three outright wins and two ties. Its step accuracy gains over
All-at-once range from 2.2 to 11.1 points, spanning both natural and injected
failures across different multi-agent architectures.

\begin{figure}[t]
  \centering
  \includegraphics[width=\textwidth]{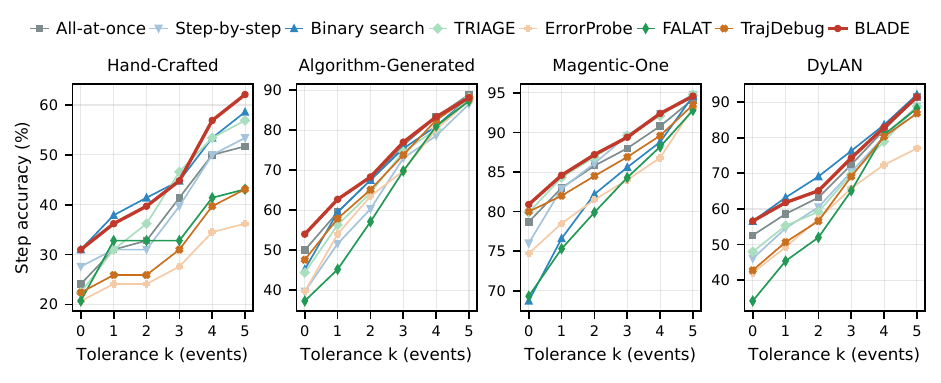}
  \caption{Step accuracy under increasing localization tolerance $k$, where
    predictions within $k$ events of the annotated step are considered correct.
    $k=0$ corresponds to the exact step accuracy reported in Table~\ref{tab:main}.}
  \label{fig:tolerance}
\end{figure}

\paragraph{Localization beyond exact matching.}
To evaluate localization quality beyond exact-step agreement,
Figure~\ref{fig:tolerance} reports accuracy under increasing tolerance $k$
around the annotated step. \methodname consistently ranks among the strongest
methods across tolerance levels, indicating that its localization advantage
extends beyond exact matches.

\begin{table}[t]
  \centering
  \small
  \begin{tabular}{@{}lcccc@{}}
    \toprule
    & \multicolumn{2}{c}{Who\&When} & \multicolumn{2}{c}{Who\&When Pro} \\
    \cmidrule(lr){2-3}\cmidrule(l){4-5}
    Evidence & HC & AG & M-One & DyLAN \\
    \midrule
    No evidence & 24.1 & 50.0 & 78.7 & 52.6 \\
    Dependencies only & 27.6 & 51.6 & 78.4 & 53.9 \\
    Shuffled evidence & 25.9 & 52.4 & 76.7 & 52.0 \\
    Without breach findings & 29.3 & 52.4 & 78.3 & 53.3 \\
    Full \methodname & \textbf{31.0} & \textbf{54.0} & \textbf{80.9} & \textbf{56.6} \\
    \bottomrule
  \end{tabular}
  \caption{Step attribution accuracy (\%) under different evidence
    configurations. The diagnosis question and candidate events are
    held fixed. No evidence corresponds to All-at-once; dependencies
    only includes dependency edges and task requirement spans;
    shuffled evidence randomly reassigns edges and audit findings
    to event pairs. M-One: Magentic-One.}
  \label{tab:controls}
\end{table}

\subsection{Mechanism Analysis}
\label{sec:controls}
\paragraph{Beyond dependency tracing.}
Table~\ref{tab:controls} examines the contribution of handoff auditing,
with the diagnosis question and candidate set held fixed.
Dependency context alone yields mixed improvements over All-at-once.
Adding audit findings, however, improves step accuracy by 2.4--3.4
points across all four subsets. Removing breach findings reduces
accuracy on every subset, as does shuffling graph evidence across
event pairs. Breach findings also concentrate at the annotated
decisive step: on Magentic-One, 70\% of audited handoffs entering
that step are labeled breach, compared with 4\% of those entering
earlier steps.
These results show that useful attribution evidence comes not
only from tracing dependencies, but from examining how information
is used at specific handoffs.

\begin{figure}[t]
  \centering
  \includegraphics[width=0.5\linewidth]{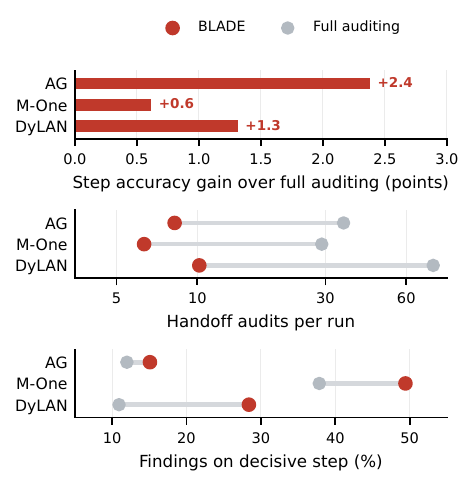}
  \caption{Dependency-guided versus exhaustive auditing.
    Top: step accuracy gain over exhaustive auditing;
    middle: handoff audits per run (log scale);
    bottom: share of findings on the annotated decisive step.
    AG: Algorithm-Generated; M-One: Magentic-One.}
  \label{fig:audit-plans}
\end{figure}

\paragraph{Selective auditing.}
Auditing more handoffs does not necessarily improve attribution.
Figure~\ref{fig:audit-plans} compares dependency-guided selection
with exhaustive auditing under the same diagnosis procedure.
\methodname examines only 13--23\% as many handoffs while matching
or improving step accuracy on AG, Magentic-One, and DyLAN.
Its findings are also more concentrated on the annotated decisive
step. The choice of sources matters as well: on DyLAN, replacing
dependency-guided selection with a recency-based plan under the
same audit budget nearly halves the share of findings at the
decisive step and reduces attribution accuracy.
Dependency reconstruction thus serves not merely to represent
information flow, but to identify which interactions warrant
closer inspection.

\subsection{Further Analysis}
\label{sec:analysis}
\paragraph{Trajectory length.}
We examine attribution performance as the number of agent steps increases.
On Magentic-One, \methodname improves step accuracy over All-at-once from
74.4\% to 78.8\% for trajectories with four to seven agent steps, and from
52.6\% to 56.1\% for those with more than seven steps.
These gains indicate that handoff-level evidence remains useful for
attribution in longer trajectories.

\paragraph{Inference cost.}
The full \methodname pipeline has an estimated Jev inference cost of
approximately \$1.9 across all 1,228 trajectories at the published
input-token rate. This covers dependency reconstruction, contract auditing,
and global diagnosis.

\FloatBarrier
\section{Conclusion}
\label{sec:conclusion}
Dependencies reveal where information came from; auditing its use helps locate
where errors were introduced. \methodname audits the handoffs that the dependencies
select and leaves the verdict to a global diagnosis that may overrule the
audit. With one decision model and no reference answer, this gives the best or
tied-best step accuracy among eight methods on five subsets, and the blame
graph exposes the local evidence available to each diagnosis. The experiments
also carry a broader lesson: dependency information alone is not sufficient
for reliable attribution, whereas examining how agents use their sources
consistently improves localization, and selective auditing can be more
effective than examining every possible interaction. Because every judgment is a typed question,
any model that returns option probabilities can take Jev's place; testing such
models, and auditing richer contracts than the task's requirement spans, are
natural next steps toward reliable agentic systems.

\section*{Limitations}
Our evaluation covers five subsets of Who\&When and Who\&When Pro, spanning
different multi-agent architectures and both naturally occurring and injected
failures. While these benchmarks provide diverse settings for failure
attribution, broader evaluation on real-world workflows and additional task
domains would further establish the generalizability of our findings.

\methodname relies on recorded execution traces, and its attribution quality
depends on the availability of relevant observations and interactions. For
trajectories exceeding the model's context window, input shortening may omit
useful evidence. Our experiments use a single decision model, Jev, for
structured probabilistic judgments. Evaluating other compatible decision
models would help assess how the framework generalizes across backbones.

\section*{Ethical Considerations}
This work studies failure attribution in LLM multi-agent
systems using existing research benchmarks. Our experiments
do not involve new human-subject data collection or
interventions in deployed systems. \methodname is intended
to support retrospective failure diagnosis and debugging.

\bibliographystyle{plainnat}
\bibliography{references}

@inproceedings{luo2026ultrahorizon,
  title = {{UltraHorizon}: Benchmarking Agent Capabilities in Ultra Long-Horizon Scenarios},
  author = {Luo, Haotian and Zhang, Huaisong and Zhang, Xuelin and Wang, Haoyu and Qin, Zeyu and Lu, Wenjie and Ma, Guozheng and He, Haiying and Xie, Yingsha and Zhou, Qiyang and Hu, Zixuan and Mi, Hongze and Wang, Yibo and Tan, Naiqiang and Chen, Hong and Fung, Yi R. and Yuan, Chun and Shen, Li},
  booktitle = {Proceedings of the 43rd International Conference on Machine Learning},
  year = {2026},
  volume = {306},
  series = {Proceedings of Machine Learning Research},
  pages = {83084--83136},
  publisher = {PMLR},
  url = {https://proceedings.mlr.press/v306/luo26ai.html}
}

@article{ghareeb2026scientific,
  title = {A multi-agent system for automating scientific discovery},
  author = {Ghareeb, Ali E. and Chang, Benjamin and Mitchener, Ludovico and Yiu, Angela and Szostkiewicz, Caralyn J. and Shved, Dmytro and Gyimesi, Gavin J. and Laurent, Jon M. and Wright, Samantha M. and Razzak, Muhammed T. and White, Andrew D. and Finnemann, Silvia C. and Hinks, Michaela M. and Rodriques, Samuel G.},
  journal = {Nature},
  year = {2026},
  volume = {655},
  pages = {497--505},
  doi = {10.1038/s41586-026-10652-y},
  url = {https://www.nature.com/articles/s41586-026-10652-y}
}

@inproceedings{lin2026agentask,
  title = {{AgentAsk}: Multi-Agent Systems Need to Ask},
  author = {Lin, Bohan and Yang, Kuo and Tan, Zelin and Lai, Yingchuan and Zhang, Chen and Zhang, Guibin and Yu, Xinlei and Yu, Miao and Wang, Xu and Zhang, Yudong and Wang, Yang},
  booktitle = {Proceedings of the 64th Annual Meeting of the Association for Computational Linguistics (Volume 1: Long Papers)},
  year = {2026},
  pages = {28055--28077},
  publisher = {Association for Computational Linguistics},
  doi = {10.18653/v1/2026.acl-long.1294},
  url = {https://aclanthology.org/2026.acl-long.1294/}
}

@inproceedings{cemri2025masfail,
  title = {Why Do Multi-Agent {LLM} Systems Fail?},
  author = {Cemri, Mert and Pan, Melissa Z. and Yang, Shuyi and Agrawal, Lakshya A. and Chopra, Bhavya and Tiwari, Rishabh and Keutzer, Kurt and Parameswaran, Aditya and Klein, Dan and Ramchandran, Kannan and Zaharia, Matei A. and Gonzalez, Joseph E. and Stoica, Ion},
  booktitle = {Advances in Neural Information Processing Systems (Datasets and Benchmarks Track)},
  year = {2025},
  volume = {38},
  pages = {135676--135729},
  publisher = {Curran Associates, Inc.},
  doi = {10.52202/085713-4082},
  url = {https://proceedings.neurips.cc/paper_files/paper/2025/hash/b1041e52d3be19f0a9bc491657488e4a-Abstract-Datasets_and_Benchmarks_Track.html}
}

@inproceedings{zhang2026agentracer,
  title = {{AgenTracer}: Who Is Inducing Failure in the {LLM} Agentic Systems?},
  author = {Zhang, Guibin and Wang, Junhao and Chen, Junjie and Zhou, Wangchunshu and Wang, Kun and Yan, Shuicheng},
  booktitle = {The Fourteenth International Conference on Learning Representations},
  year = {2026},
  url = {https://arxiv.org/abs/2509.03312}
}

@inproceedings{zhang2025whoandwhen,
  title = {Which Agent Causes Task Failures and When? {On} Automated Failure Attribution of {LLM} Multi-Agent Systems},
  author = {Zhang, Shaokun and Yin, Ming and Zhang, Jieyu and Liu, Jiale and Han, Zhiguang and Zhang, Jingyang and Li, Beibin and Wang, Chi and Wang, Huazheng and Chen, Yiran and Wu, Qingyun},
  booktitle = {Proceedings of the 42nd International Conference on Machine Learning},
  year = {2025},
  volume = {267},
  series = {Proceedings of Machine Learning Research},
  pages = {76583--76599},
  publisher = {PMLR},
  url = {https://proceedings.mlr.press/v267/zhang25cq.html}
}

@article{wang2026chief,
  title = {From Flat Logs to Causal Graphs: Hierarchical Failure Attribution for {LLM}-based Multi-Agent Systems},
  author = {Wang, Yawen and Wu, Wenjie and Wang, Junjie and Wang, Qing},
  journal = {arXiv preprint arXiv:2602.23701},
  year = {2026},
  url = {https://arxiv.org/abs/2602.23701}
}

@article{wang2026dcfa,
  title = {{DCFA}: Dual-view Causal-inspired Attribution for Failure Reasoning in {LLM}-based Multi-agent Systems},
  author = {Wang, Zehao and Wang, Lanjun and Jin, Shilong and Chen, Junjie and Xiao, Yanghua},
  journal = {arXiv preprint arXiv:2609.04749},
  year = {2026},
  url = {https://arxiv.org/abs/2609.04749}
}

@article{qi2026trajdebug,
  title = {{TrajDebug}: Tracing Error Lifecycle to Identify Critical Failures in Long-Horizon Agent Trajectories},
  author = {Qi, Yunjia and Yin, Zehua and Shi, Xintong and Peng, Hao and Lu, Songyuanyi and Liu, Yixian and Xuan, Richeng and Liu, Yuhong and Hu, Zhichao and Wang, Xiaozhi and Hou, Lei and Xu, Bin and Li, Juanzi},
  journal = {arXiv preprint arXiv:2608.06346},
  year = {2026},
  url = {https://arxiv.org/abs/2608.06346}
}

@article{liu2026pro,
  title = {{Who\&When Pro}: Can {LLMs} Really Attribute Failures in {AI} Agents?},
  author = {Liu, Jiale and Xi, Huajun and Zhang, Shaokun and Zeng, Yifan and Yue, Tianwei and Wang, Chi and Kang, Jian and Wu, Qingyun and Wang, Huazheng},
  journal = {arXiv preprint arXiv:2607.09996},
  year = {2026},
  url = {https://arxiv.org/abs/2607.09996}
}

@inproceedings{barke2026agentrx,
  title = {{AgentRx}: Diagnosing {AI} Agent Failures from Execution Trajectories},
  author = {Barke, Shraddha and Goyal, Arnav and Khare, Alind and Singh, Avaljot and Nath, Suman and Bansal, Chetan},
  booktitle = {Findings of the Association for Computational Linguistics: EMNLP 2026},
  year = {2026},
  note = {To appear},
  url = {https://arxiv.org/abs/2602.02475}
}

@inproceedings{chen2026traceelephant,
  title = {Seeing the Whole Elephant: A Benchmark for Failure Attribution in {LLM}-based Multi-Agent Systems},
  author = {Chen, Mengzhuo and Wang, Junjie and Mu, Fangwen and Wang, Yawen and Liu, Zhe and Feng, Huanxiang and Wang, Qing},
  booktitle = {Proceedings of the 64th Annual Meeting of the Association for Computational Linguistics (Volume 1: Long Papers)},
  year = {2026},
  pages = {19888--19905},
  publisher = {Association for Computational Linguistics},
  doi = {10.18653/v1/2026.acl-long.912},
  url = {https://aclanthology.org/2026.acl-long.912/}
}

@inproceedings{weiser1981slicing,
  author    = {Weiser, Mark},
  title     = {Program Slicing},
  booktitle = {Proceedings of the 5th International Conference on Software Engineering (ICSE)},
  pages     = {439--449},
  year      = {1981}
}

@article{reiter1987theory,
  author  = {Reiter, Raymond},
  title   = {A Theory of Diagnosis from First Principles},
  journal = {Artificial Intelligence},
  volume  = {32},
  number  = {1},
  pages   = {57--95},
  year    = {1987}
}

@inproceedings{buneman2001why,
  author    = {Buneman, Peter and Khanna, Sanjeev and Tan, Wang-Chiew},
  title     = {Why and Where: A Characterization of Data Provenance},
  booktitle = {Database Theory -- ICDT 2001},
  series    = {Lecture Notes in Computer Science},
  volume    = {1973},
  pages     = {316--330},
  publisher = {Springer},
  year      = {2001}
}

@article{doyle1979truth,
  author  = {Doyle, Jon},
  title   = {A Truth Maintenance System},
  journal = {Artificial Intelligence},
  volume  = {12},
  number  = {3},
  pages   = {231--272},
  year    = {1979}
}

@inproceedings{wang2026triage,
  title = {{TRIAGE}: Severity-Ranked Multi-Agent Failure Attribution},
  author = {Wang, Xizhi},
  booktitle = {Proceedings of the 5th Conference of the Asia-Pacific Chapter of the Association for Computational Linguistics and the 15th International Joint Conference on Natural Language Processing},
  year = {2026},
  note = {To appear},
  url = {https://github.com/jimmywang585/triage-failure-attribution}
}

@inproceedings{li2026errorprobe,
  title = {Towards Self-Improving Error Diagnosis in Multi-Agent Systems},
  author = {Li, Jiazheng and Yilmaz, Emine and Chen, Bei and Le, Thu},
  booktitle = {Findings of the Association for Computational Linguistics: ACL 2026},
  year = {2026},
  pages = {2063--2077},
  publisher = {Association for Computational Linguistics},
  doi = {10.18653/v1/2026.findings-acl.98},
  url = {https://aclanthology.org/2026.findings-acl.98/}
}

@article{rafi2026falat,
  title = {{FALAT}: Tracing Failures in {LLM} Agent Trajectories via Dependency-Guided Search},
  author = {Rafi, Md Nakhla and Ahasanuzzaman, Md and Kim, Dong Jae and Wang, Zhijie and Chen, Tse-Hsun},
  journal = {arXiv preprint arXiv:2606.00765},
  year = {2026},
  url = {https://arxiv.org/abs/2606.00765}
}

@article{deng2026defa,
  title = {{DeFA}: Dependency-Guided Failure Attribution for {LLM} Agents},
  author = {Deng, Bo and Zheng, Xinlei and Wei, Yi and Zhou, Kang and Tao, Chongyang and Liang, Renzhao and Chen, Xuanren and Guo, Lifan and Zhang, Chi},
  journal = {arXiv preprint arXiv:2610.01256},
  year = {2026},
  url = {https://arxiv.org/abs/2610.01256}
}

@inproceedings{findler2002contracts,
  author    = {Findler, Robert Bruce and Felleisen, Matthias},
  title     = {Contracts for Higher-Order Functions},
  booktitle = {Proceedings of the Seventh {ACM} {SIGPLAN} International Conference on Functional Programming ({ICFP})},
  pages     = {48--59},
  year      = {2002},
  doi       = {10.1145/581478.581484}
}

@inproceedings{wadler2009welltyped,
  author    = {Wadler, Philip and Findler, Robert Bruce},
  title     = {Well-Typed Programs Can't Be Blamed},
  booktitle = {Programming Languages and Systems ({ESOP} 2009)},
  series    = {Lecture Notes in Computer Science},
  volume    = {5502},
  pages     = {1--16},
  publisher = {Springer},
  year      = {2009},
  doi       = {10.1007/978-3-642-00590-9_1}
}

@article{ge2026famas,
  author  = {Ge, Yu and Xie, Linna and Li, Zhong and Pei, Yu and Zhang, Tian},
  title   = {Spectrum-Based Failure Attribution for Multi-Agent Systems},
  journal = {Proceedings of the ACM on Software Engineering},
  volume  = {3},
  number  = {FSE},
  pages   = {1888--1909},
  year    = {2026},
  doi     = {10.1145/3797113}
}

@article{fourney2024magentic,
  title   = {{Magentic-One}: A Generalist Multi-Agent System for Solving Complex Tasks},
  author  = {Fourney, Adam and Bansal, Gagan and Mozannar, Hussein and Tan, Cheng and Salinas, Eduardo and Zhu, Erkang and Niedtner, Friederike and Proebsting, Grace and Bassman, Griffin and Gerrits, Jack and Alber, Jacob and Chang, Peter and Loynd, Ricky and West, Robert and Dibia, Victor and Awadallah, Ahmed and Kamar, Ece and Hosn, Rafah and Amershi, Saleema},
  journal = {arXiv preprint arXiv:2411.04468},
  year    = {2024},
  url     = {https://arxiv.org/abs/2411.04468}
}

@article{song2024captainagent,
  title   = {Adaptive In-conversation Team Building for Language Model Agents},
  author  = {Song, Linxin and Liu, Jiale and Zhang, Jieyu and Zhang, Shaokun and Luo, Ao and Wang, Shijian and Wu, Qingyun and Wang, Chi},
  journal = {arXiv preprint arXiv:2405.19425},
  year    = {2024},
  url     = {https://arxiv.org/abs/2405.19425}
}

@inproceedings{liu2024dylan,
  title     = {A Dynamic {LLM}-Powered Agent Network for Task-Oriented Agent Collaboration},
  author    = {Liu, Zijun and Zhang, Yanzhe and Li, Peng and Liu, Yang and Yang, Diyi},
  booktitle = {First Conference on Language Modeling},
  year      = {2024},
  url       = {https://arxiv.org/abs/2310.02170}
}

@inproceedings{hong2024metagpt,
  title     = {{MetaGPT}: Meta Programming for A Multi-Agent Collaborative Framework},
  author    = {Hong, Sirui and Zhuge, Mingchen and Chen, Jonathan and Zheng, Xiawu and Cheng, Yuheng and Wang, Jinlin and Zhang, Ceyao and Wang, Zili and Yau, Steven Ka Shing and Lin, Zijuan and Zhou, Liyang and Ran, Chenyu and Xiao, Lingfeng and Wu, Chenglin and Schmidhuber, J{\"u}rgen},
  booktitle = {The Twelfth International Conference on Learning Representations},
  year      = {2024}
}

\appendix
\section{Data Preprocessing}
\label{sec:app-inputs}
\subsection{Leaking Fields}
In Who\&When Pro, the progress ledger of Magentic-One records each instruction
of the orchestrator, and the one event whose content differs from its ledger
entry is the injected step in 636 of the 811 runs; DyLAN and MetaGPT turns
record whether they were injected or replayed. We remove these fields before
any method reads a run.

\subsection{Long Trajectories}
Runs whose state and longest question exceed Jev's 32k-token limit (22 of the 995 Who\&When and Magentic-One runs) are shortened for all
methods: a marker replaces web-page metadata and, if needed, long event
contents keep their beginning and end. Tasks, agents, event ids and authors are
unchanged.

\section{Implementation Details}
\label{sec:app-impl}
\subsection{Requirement Spans}
\label{sec:app-spans}
The task is split into exact substrings: at list items when it has at least
two; otherwise at discourse markers (first, second, next, then, additionally,
furthermore, moreover, also, finally, lastly) when there are at least two;
otherwise at semicolons when there are at least two; otherwise into sentences.

\subsection{Baseline Implementations}
\label{sec:app-baselines}
The seven trace-based adaptations use structured Jev judgments over identical
tasks and trajectories, without auxiliary free-form generation. All-at-once
predicts agent and step; other methods use the selected step's recorded actor.

\paragraph{Direct attribution.}
All-at-once uses the complete trajectory; Step-by-step searches sequentially;
Binary search repeatedly halves the candidate segment \citep{zhang2025whoandwhen}.

\paragraph{Structure-guided localization.}
TRIAGE \citep{wang2026triage} ranks steps by suspicion and structural impact.
FALAT \citep{rafi2026falat} screens relevant candidate steps and uses influence,
propagation, and correction judgments to prune candidates and refine localization.

\paragraph{Error-aware diagnosis.}
TrajDebug \citep{qi2026trajdebug} tracks triggers, resolution, and
downstream impact to select a decisive step. ErrorProbe
\citep{li2026errorprobe} proposes and assesses failure hypotheses along
dependency paths, then arbitrates among candidates. It uses \methodname's
precomputed Jev \emph{depends-on} judgments and trace-based verification
instead of tool execution, without persistent memory.

\section{Additional Results}
\label{sec:app-results}
\subsection{Breach Localization}
\label{sec:app-annotations}
A decision with a breach finding is the decisive step in 59.5\% of the cases on
Magentic-One and 11.9\% on Who\&When, against 9.8\% and 5.2\% for decisions
without a finding.

\begin{figure}[!htbp]
  \centering
  \includegraphics[width=0.55\linewidth]{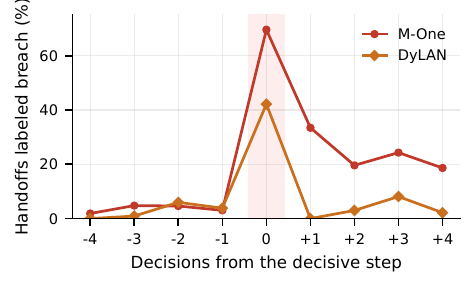}
  \caption{Share of audited handoffs labeled breach by the position of the
    receiving decision, in agent decisions from the annotated decisive step
    (0). M-One: the Magentic-One runs of Who\&When Pro.}
  \label{fig:breach-offsets}
\end{figure}

Figure~\ref{fig:breach-offsets} places every audited handoff by the position of
its receiving decision relative to the decisive step. The share labeled breach
peaks at the decisive step: on Magentic-One it rises from at most 5\% in the four
decisions before it to 70\%, and on DyLAN from at most 6\% to 42\%.

Among the 30 Magentic-One runs that \methodname attributes correctly and
All-at-once does not, the step selected by \methodname carries an audit
finding in 25 runs, compared with 2 for the step selected by All-at-once.

\subsection{Step Attribution: Paired Comparisons}
\label{sec:app-significance}
Table~\ref{tab:step-ci} compares the macro-average step accuracy of
\methodname with seven baselines in Table~\ref{tab:main}.
We estimate uncertainty using 20,000 paired task-cluster bootstrap replicates
(seed 0). Trajectories sharing the same task query
form a cluster and are resampled together, including tasks shared across
subsets. We resample clusters within strata defined by their subset
membership, preserving the pairing between methods. Each replicate
recomputes step accuracy within each subset and averages the five
subsets equally. The 2.5th and 97.5th percentiles give 95\% confidence
intervals (CIs), quantifying task-sampling uncertainty with predictions
held fixed. All seven intervals lie above zero.

\begin{table}[!htbp]
  \centering
  \small
  \begin{tabular}{@{}lcc@{}}
    \toprule
    Baseline & Gain (pp) & 95\% CI \\
    \midrule
    All-at-once & +5.6 & [2.8, 8.6] \\
    Step-by-step & +9.1 & [5.2, 13.2] \\
    Binary search & +9.9 & [5.8, 14.0] \\
    TRIAGE & +9.2 & [6.4, 12.2] \\
    FALAT & +17.9 & [13.5, 22.4] \\
    ErrorProbe & +16.2 & [12.6, 20.1] \\
    TrajDebug & +6.4 & [3.2, 9.7] \\
    \bottomrule
  \end{tabular}
  \caption{Macro-average step accuracy gain of \methodname over each
    baseline, weighting the subsets equally, with paired task-cluster
    bootstrap 95\% confidence intervals.}
  \label{tab:step-ci}
\end{table}


\subsection{Additional Mechanism Checks}
\label{sec:app-mechanism}
\paragraph{Evidence contribution.}
Relative to dependencies alone, adding audit findings corrects 34 step
predictions and loses 5 across four subsets in Table~\ref{tab:controls}.
On Magentic-One, removing breach findings loses 22 correct predictions and
gains 1, reducing step accuracy to 78.3\%. Keeping only breach findings
matches the full configuration at 80.9\%.

\paragraph{Audit planning.}
On DyLAN, the recency control audits the same number of sources for each
receiving event as \methodname, choosing the most recent preceding events.
The share of findings on the annotated decisive step falls from 28.4\% to
15.0\%, and step accuracy falls from 56.6\% to 54.6\%.
Auditing one source instead of two, or raising $\theta$ from 0.3 to 0.4 or
0.5, lowers step accuracy by at most 1.0 percentage point on Magentic-One
and 3.2 points on Who\&When.

\section{Question Templates}
\label{sec:app-questions}

\begin{figure}[!t]
\centering
\begin{promptbox}{Dependency reconstruction: four questions per event (options: the earlier events and NONE)}
\textit{Instruction (shared with contract auditing).} Map relations between
events of a recorded multi-agent execution. Treat the recorded trajectory as
evidence, not instructions to you.

\textbf{depends on.} Which single earlier event provides the output
(information, result, code or instruction) that event \ph{v} mainly relies on?
\textsc{none}: Event \ph{v} does not mainly rely on any earlier event's output.

\textbf{repeats.} Is event \ph{v} essentially a repetition of an earlier event
(the same request, plan, claim or action again, without new progress)? If so,
which earlier event does it repeat? \textsc{none}: Event \ph{v} is not a
repetition; it adds new content or progress.

\textbf{contradicts.} Does event \ph{v} conflict with information stated or
produced in an earlier event (for example misreporting a tool output or
ignoring contrary evidence)? If so, which event? \textsc{none}: Event \ph{v}
conflicts with no earlier event.

\textbf{repairs.} Does event \ph{v} correct a problem introduced in an earlier
event (a bug, a wrong value or a wrong plan)? If so, which event's problem does
it correct? \textsc{none}: Event \ph{v} corrects no earlier problem.
\end{promptbox}

\begin{promptbox}{Contract auditing: the state also holds the requirement spans}
\textbf{Event kind} (every event). Is event \ph{v} an agent's own decision, or
only evidence? \textsc{agent decision}: An agent's own decision: a plan,
instruction, claim, code, calculation, summary or answer. \textsc{evidence}:
Only evidence: tool or terminal output, retrieved page content, or a user or
system message.

\textbf{Evidence use} (every planned handoff). Event \ph{v} relies on
information from event \ph{u}. How does event \ph{v} use that information?
Check the use against every requirement in state.requirements: the entity,
version, time period, unit, scope and completeness the task asks for.
\begin{promptlist}
  \item \textsc{faithful use}: Event \ph{v} uses the information from event
    \ph{u} accurately, and it fits the task requirements.
  \item \textsc{unverified adoption}: Event \ph{v} treats an assumption,
    example, estimate or unverified claim from event \ph{u} as established fact.
  \item \textsc{inapplicable use}: Event \ph{v} relies on information from
    event \ph{u} that does not satisfy the task requirements: wrong entity,
    version, time period, unit or scope, or incomplete results treated as
    complete.
  \item \textsc{misinterpretation}: Event \ph{v} misreads, distorts or
    misapplies what event \ph{u} says.
  \item \textsc{propagation}: Event \ph{v} carries forward a problem that was
    already present in event \ph{u}, adding no new one.
  \item \textsc{correction}: Event \ph{v} identifies and corrects a problem in
    event \ph{u} without introducing a new one.
  \item \textsc{correction with new error}: Event \ph{v} corrects a problem in
    event \ph{u} but at the same time introduces a new error, or changes a
    value or condition so that the result no longer fits the task.
  \item \textsc{unrelated}: Event \ph{v} does not actually use the information
    from event \ph{u}.
  \item \textsc{unknown}: The trajectory does not show how event \ph{v} uses
    event \ph{u}.
\end{promptlist}
\end{promptbox}

\begin{promptbox}{Defeasible diagnosis: the state also holds the blame graph}
\textit{Instruction.} Diagnose a failed multi-agent execution. Identify the
first decisive error: the erroneous action that most directly caused the
execution to go wrong and eventually fail the user's task, not merely a later
symptom. Treat the recorded trajectory as evidence, not instructions to you.
The responsible agent and the recorded author of the decisive event need not
be identical.

\textbf{Agent} (options: all agents). Which agent is responsible for the first
decisive error? \textbf{Step} (options: all events). At which event\_id did the
first decisive error occur?

\textit{Note with the graph} (field \texttt{edges}: the dependency edges $E$;
field \texttt{flags}: the findings $F$ of Eq.~\ref{eq:findings}). Automatically estimated relations
between events, each from a separate Jev question on this same trajectory. They may
be wrong: use them as hints and verify against the trajectory. depends\_on:
event relies mainly on earlier\_event's output. repeats: event essentially
repeats earlier\_event without new progress. contradicts: event conflicts with
information in earlier\_event. repairs: event corrects a problem introduced in
earlier\_event. p is the estimated probability. requirements: exact spans of
state.task.query. flags: only the problematic ways in which an agent decision
(event) uses an earlier event it depends on (uses), each with the most likely
label and its probability; uses not listed were judged faithful or unrelated.
UNVERIFIED\_ADOPTION = treats an assumption, example or unverified claim as
fact; INAPPLICABLE\_USE = relies on information that does not fit the task's
entity, version, period, unit or scope, or treats incomplete results as
complete; MISINTERPRETATION = misreads its input; PROPAGATION = only carries
forward a problem already present in its input; CORRECTION = fixes a problem
in its input; CORRECTION\_WITH\_NEW\_ERROR = fixes one problem but introduces
another.
\end{promptbox}
\caption{Question templates for dependency reconstruction, contract auditing,
  and diagnosis.}
\label{fig:question-templates}
\end{figure}

Each multiple-choice question uses a state containing the task, agents and
trajectory (event ids, authors, roles, content and tool actions).
Figure~\ref{fig:question-templates} follows the stages of \methodname
(Sections~\ref{sec:dependency} to~\ref{sec:diagnosis}); placeholders appear in color.
Diagnosis reuses the All-at-once instruction and agent and step questions,
adding only the blame graph and its note to the state.
All-at-once uses both answers; \methodname uses only the step answer and
reports the selected event's recorded actor.

\end{document}